\documentclass[letterpaper, 10 pt, conference]{ieeeconf}

\IEEEoverridecommandlockouts

\usepackage{cite}
\usepackage{amsmath,amssymb,amsfonts}
\usepackage{algorithmic}
\usepackage{graphicx}
\usepackage{textcomp}
\usepackage{xcolor}
\usepackage{siunitx}
\usepackage{float}
\usepackage[symbol]{footmisc}
\usepackage{ar}
\usepackage{url}
\usepackage{balance}

\newcommand{\bs}[1]{\boldsymbol{#1}}  
\newcommand{\ts}[1]{\text{#1}} 

\renewcommand{\baselinestretch}{0.91}

\def\BibTeX{{\rm B\kern-.05em{\sc i\kern-.025em b}\kern-.08em

    T\kern-.1667em\lower.7ex\hbox{E}\kern-.125emX}}

\begin{document}

\title{\LARGE \bf A New \mbox{10-mg} SMA-Based Fast Bimorph Actuator for Microrobotics\\

\thanks{This work was partially funded by the Washington State University (WSU) Foundation and the Palouse Club through a Cougar Cage Award to \mbox{N.\,O.\,P\'erez-Arancibia}. Additional funding was provided by the WSU Voiland College of Engineering and Architecture through a start-up fund to \mbox{N.\,O.\,P\'erez-Arancibia}.} %
\thanks{The authors are with the School of Mechanical and Materials Engineering, Washington State University (WSU), Pullman,\,WA\,99164,\,USA. Corresponding authors' \mbox{e-mail:} {\tt conor.trygstad@wsu.edu}~(C.\,K.\,T.); {\tt n.perezarancibia@wsu.edu} (N.\,O.\,P.-A.).}%
}
\author{Conor K. Trygstad, Elijah K. Blankenship, and N\'estor O. P\'erez-Arancibia}

\maketitle
\thispagestyle{empty}
\pagestyle{empty}

\begin{abstract}
We present a new \mbox{millimeter-scale} bimorph actuator for microrobotic applications, driven by feedforward controlled \textit{\mbox{shape-memory} alloy} (SMA) wires. The device weighs $\bs{10}\,\ts{mg}$, measures $\bs{14}\,\ts{mm}$ in length, and occupies a volume of $\bs{4.8}\,\ts{mm}^{\bs{3}}$, which makes it the lightest and smallest fully functional \mbox{SMA-based} bimorph actuator for microrobotics developed to date. The experimentally measured operational bandwidth is on the order of $\bs{20}\,\ts{Hz}$, and the unimorph and bimorph maximum \mbox{low-frequency} displacement outputs are on the order of \mbox{$\bs{3.5}$} and \mbox{$\bs{7}\,\ts{mm}$}, respectively. To test and demonstrate the functionality and suitability of the actuator for microrobotics, we developed the \textit{\mbox{Fish-\&-Ribbon--Inspired} Small Swimming Harmonic roBot} (\mbox{FRISSHBot}). Loosely inspired by \textit{carangiformes}, the \mbox{FRISSHBot} leverages \textit{fluid-structure interaction}~(FSI) phenomena to propel itself forward, weighs $\bs{30}\,\ts{mg}$, measures $\bs{34}\,\ts{mm}$ in length, operates at frequencies of up to $\bs{4}\,\ts{Hz}$, and swims at speeds of up to \mbox{$\bs{3.06}\,\ts{mm}\cdot\ts{s}^{\bs{-1}}$}~(\mbox{$\bs{0.09}\,\ts{Bl}\cdot\ts{s}^{\bs{-1}}$}). This robot is the lightest and smallest swimmer with onboard actuation developed to date.
\end{abstract}

\section{Introduction} 
\label{SEC01}
\vspace{-0.5ex}
The vision of \mbox{insect-scale} robotic swarms working in harmony with humans to complete essential tasks for society will become a reality only once critical challenges in microfabrication, sensing, actuation, power, and computation are solved. One of these challenges is the creation of lightweight microactuators with low power consumption and versatile functionality. Numerous advanced and novel \mbox{mm-to-cm--scale} microsystems have been developed during the past few years using predominantly piezoelectric~\cite{song2007PZTWaterStrider, lee2011design,goldberg2018power,wu2019insect,gravish2020stcrawler,BeePlus_2019,BeePlusPlus_2023,Li2023PZT}, electromagnetic~\cite{contreras2017first, lu2018bioinspired, hu2018small, pierre20183d}, \mbox{\textit{dielectric--elastomer}~(DE)}~\cite{Duduta2017DEA, ren2022DEAFlyer, Berlinger2018DEAFish, Ji2019DEAnsect}, rotational motor~\cite{MilliMobile2023,hutama2021UltrasonicMotor,Truong2019JumpingRobot,Chen2020BioinspiredFlapper}, and \textit{\mbox{shape-memory} alloy}~(SMA)~\cite{SMALLBug_2020, SMARTI_2021, RoBeetle_2020, Waterstrider_2023, VLEIBot_2024} actuation technologies. While, in the aggregate, these results represent innovation and progress in microrobotic design, rapid prototyping, control performance, autonomy, and energy efficiency, all the platforms presented in~\cite{song2007PZTWaterStrider, lee2011design,goldberg2018power,wu2019insect,gravish2020stcrawler,BeePlus_2019,BeePlusPlus_2023,Li2023PZT,contreras2017first,lu2018bioinspired, hu2018small,pierre20183d,Duduta2017DEA, ren2022DEAFlyer, Berlinger2018DEAFish, Ji2019DEAnsect,MilliMobile2023,hutama2021UltrasonicMotor,Truong2019JumpingRobot,Chen2020BioinspiredFlapper} are limited by the need for complex electronics and lack of sources of power with high energy densities. For obvious reasons, microactuators that require low operational power and simple electronics, generate \mbox{high-force} outputs, and exhibit high versatility are a superior choice for advanced autonomous microrobotics. One promising technological path in this direction is \mbox{SMA-based} actuation of the type presented in~\cite{SMALLBug_2020, SMARTI_2021, RoBeetle_2020, Waterstrider_2023, VLEIBot_2024}, which exhibits \textit{\mbox{high-work} densities} (HWD) and requires low voltages of operation---typically, $1$ to \mbox{$25\,\ts{V}$}. Because of this second characteristic, these \mbox{SMA-based} actuators can be driven using relatively simple power electronics. Furthermore, they can be excited through exothermic chemical reactions using, for example, \mbox{Pt-based} catalytic combustion of hydrogen, methanol, or hydrocarbons, which is highly advantageous from the power and endurance perspectives due to the unparalleled specific energies exhibited by these fuels when compared to the best \mbox{state-of-the-art} batteries commercially available~\cite{RoBeetle_2020}. 
\begin{figure}[t!]
\vspace{1.5ex}
\begin{center}
\includegraphics[width=0.48\textwidth]{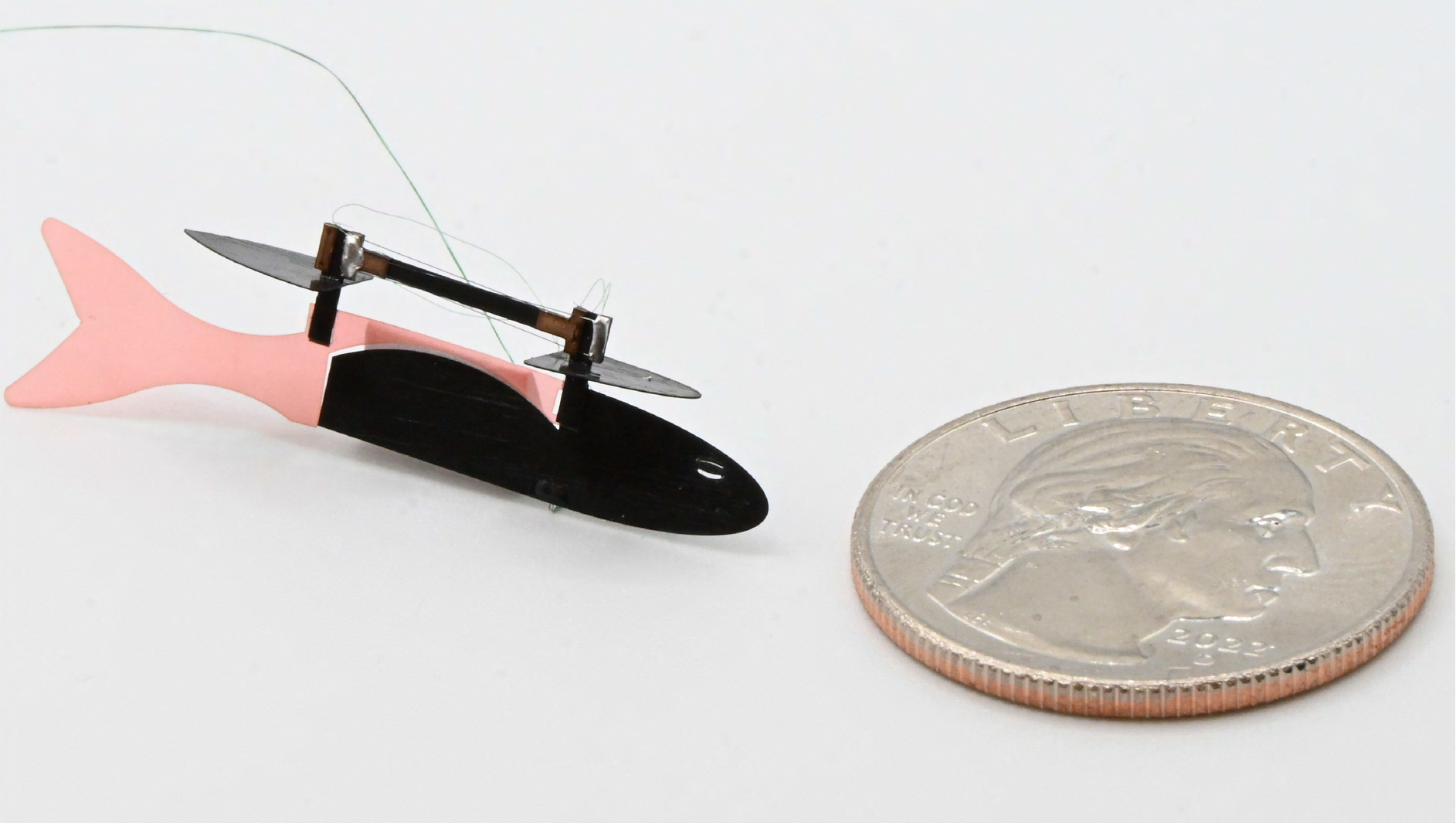}
\end{center}
\vspace{-2.0ex}    
\caption{\textbf{Photograph of the \textit{Fish-\&-Ribbon--Inspired Small Swimming Harmonic roBot} (FRISSHBot).} The FRISSHBot is a \mbox{$30$-mg} microswimmer driven by the new \mbox{$10$-mg} \mbox{SMA-based} fast bimorph actuator introduced in this paper. \label{FIG01}}
\vspace{-2.0ex}
\end{figure}

The recent creation of advanced \mbox{high-performance} crawling and swimming microrobots---such as the \mbox{MiniBug}, \mbox{WaterStrider}, and VLEIBot---was enabled by the development of the fully functional HWD \mbox{SMA-based} microactuators presented in~\cite{SMALLBug_2020, SMARTI_2021, RoBeetle_2020, Waterstrider_2023, VLEIBot_2024}, which, with weights in the range of $1$ to \mbox{$10$\,mg}, are the lightest actuators of this type reported to date. These devices have been demonstrated to operate at frequencies up to $40\,\ts{Hz}$ and produce forces up to \mbox{$84\,\ts{mN}$}, thus enabling both high performance and energy efficiency at the \mbox{mm-to-cm--scale}. Despite their excellent functionality and performance, these prior microactuators have a unimorph configuration, which greatly limits their applicability in microrobotic systems that require symmetric actuation patterns. To address this issue, here we introduce a new \mbox{$10$-mg} HWD \mbox{SMA-based} bimorph microactuator capable of achieving both high bidirectional frequencies of operation and large displacement outputs, when excited with \textit{\mbox{pulse-width-modulation}} (PWM) voltages. Through dynamic characterization experiments, we determined that the proposed actuator can function bidirectionally at frequencies as high as $20\,\ts{Hz}$ and can generate a \textit{maximum actuator displacement output} (MADO) on the order of $3.5$ and \mbox{$7\,\ts{mm}$} in \mbox{low-frequency} (\mbox{$\sim \hspace{-0.3ex} 1$\,Hz}) unimorph and bimorph operation, respectively. Although the approach presented here is novel, the notion of \mbox{SMA-based} bimorph actuation is not entirely original. For example, \mbox{film-based} bimorph SMA actuators are described in~\cite{Knick2019MEMSBimorph, Sun2020Liftoff, Seigner2021OrigamiBimorph}. While these actuators are the result of highly innovative and sophisticated fabrication techniques, which have opened new avenues for the integration of microsystems in small robotic platforms, \mbox{high-frequency} \mbox{($> \hspace{-0.6ex} 1\,\ts{Hz}$)} operation of these devices has not yet been reported. The inability to operate at frequencies higher than $1\,\ts{Hz}$ greatly limits the functionality of an actuator, especially in microrobotic propulsion. Other types of \mbox{SMA-based} bimorph actuators are presented in~\cite{Geetha2022BimorphSwitch,Chen2022BendingSMA,Kennedy2020BimorphSMA}. These devices can be fabricated with dimensions at the \mbox{cm-scale} and show great promise for integration in robotic systems that employ \mbox{real-time} feedback loops during operation. However, their geometrical designs are significantly larger than that of the actuator introduced here---greater than $50\,\ts{mm}^3$ versus $4.8\,\ts{mm}^3$ by volume---and, additionally, their implementation into robotic platforms has not yet been reported.

The main motivation that prompted us to develop new \mbox{mm-to-cm--scale} \mbox{high-frequency} \mbox{($> \hspace{-0.6ex} 1\,\ts{Hz}$)} \mbox{SMA-based} bimorph microactuators is the necessity to increase the maneuverability and controllability of microrobotic swimmers such as the \mbox{WaterStrider}, \mbox{VLEIBot}, and \mbox{VLEIBot\textsuperscript{+}} presented in~\cite{Waterstrider_2023} and \cite{VLEIBot_2024}. Also, we hypothesize that other aquatic microrobots with propulsors that employ undulatory motions to generate forward thrust---e.g., those described in~\cite{Cho2008SMARibbon}~and~\cite{Rossi2011BendingSMA}---can greatly benefit from the technological approach introduced here. Due to their unimorph design, the \mbox{SMA-based} actuators that propel the bioinspired microswimmers in~\mbox{\cite{Waterstrider_2023}~and~\cite{VLEIBot_2024}} produce asymmetric forces that highly depend on the direction of actuation. This dynamic characteristic significantly diminishes the locomotion performance and maneuverability achieved by the \mbox{VLEIBot} swimmer, also affecting its experimental controllability. To solve this problem, we introduced the \mbox{dual-propulsor} configuration that drives the \mbox{VLEIBot\textsuperscript{+}}. However, while achieving high maneuverability, the measured locomotion performance, in terms of forward speed, of this swimmer is highly sensitive to a \mbox{force-balance} tuning of the transmission mechanism that maps the actuator output to the undulatory motions of the propulsors that generate swimming. In the cases of the \mbox{VLEIBot} and \mbox{VLEIBot\textsuperscript{+}}~\cite{VLEIBot_2024}, the use of bimorph actuation would enable frequency modulation of the propulsors' tails without the need for offline tuning; in the case of the WaterStrider~\cite{Waterstrider_2023}, the use of bimorph actuation would enable bidirectional locomotion capabilities. These examples highlight the rationale and potential utility of developing new HWD small and lightweight \mbox{SMA-based} bimorph actuators for microrobotics.
\begin{figure}[t!]
\vspace{1.5ex}    
\begin{center}    
\includegraphics[width=0.48\textwidth]{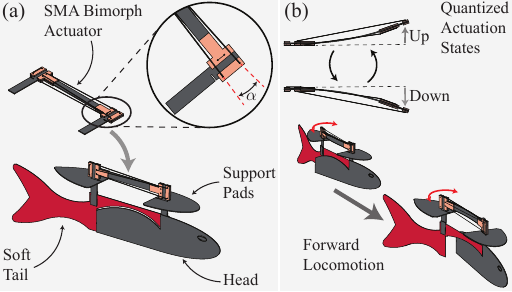}
\end{center}
\vspace{-2ex}
\caption{\textbf{Robotic design, actuation functionality, and locomotion}.
\textbf{(a)}~The \mbox{FRISSHBot} is composed of four main components: (i)~an \mbox{SMA-based} bimorph actuator, installed with integrated fabrication features, that drives the swimmer's propulsor; (ii)~two pads made of CF that support the weight of the swimmer on water using surface tension; (iii)~a rigid head that, interacting with water, generates high lateral drag relative to that produced by the swimmer's tail; and, (iv)~a soft tail made of \mbox{$25$-{\textmu}m-thick} fluoropolymer film~\mbox{(AirTech~A4000R14417)} and with a shape bioinspired by \mbox{forked-caudal} fins. We hypothesize that the rigid head, by producing high lateral drag, \textit{anchors} the body of the swimmer while the tail undulates in the surrounding fluid. The SMA wires that drive each subsystem of the bimorph actuator are installed symmetrically with respect to the device's longitudinal axis with an angle $\alpha$ (see inset). This design feature prevents the relaxed SMA wires at one side of the actuator from colliding with the central CF beam of the device's structure when the SMA wires at the other side contract due to Joule heating. \textbf{(b)}~Functionality of the proposed \mbox{SMA-based} bimorph actuator and microswimmer. The actuator was conceived to continuously transition between the two possible directions of deformation---up and down quantized states---corresponding to the sequential contraction and relaxation of the SMA subsystems at each side of the device. During locomotion, the bimorph actuator oscillates the swimmer's soft tail with large amplitudes relative to those excited for the rigid head. We hypothesize that hydrodynamic interactions between the flows generated by the \mbox{FRISSHBot's} head and tail produce the thrust required for forward propulsion. \label{FIG02}}
\vspace{-2ex}
\end{figure}

With a weight of \mbox{$10$\,mg}, length of \mbox{$14$\,mm}, and volume of \mbox{$4.8\,\ts{mm}^3$}, the device presented in this paper is the lightest and smallest fully functional \mbox{SMA-based} bimorph actuator for microrobotics reported to date. To test and demonstrate its functionality and suitability for driving microrobots, we designed and fabricated the \textit{\mbox{Fish-\&-Ribbon--Inspired} Small Swimming Harmonic roBot} (FRISSHBot), shown in~Fig.\,\ref{FIG01}. This microswimmer leverages the relatively strong surface tension of water to stay afloat. We embedded the proposed \mbox{SMA-based} bimorph actuator into the FRISSHBot's structure using integrated geometrical features that facilitate fabrication. As a result, the developed prototype is extremely lightweight (\mbox{$30\,\ts{mg}$}), has a simple design, and achieves high performance in terms of forward speed for its weight. Namely, the \mbox{FRISSHBot} can operate at frequencies of up to $4\,\ts{Hz}$ and locomote at speeds as high as \mbox{$3.06\,\ts{mm}\cdot\ts{s}^{-1}$} (\mbox{$0.09\,\ts{Bl}\cdot\ts{s}^{-1}$}). Overall, the experimental results obtained with this platform compellingly demonstrate the potential of the proposed \mbox{SMA-based} bimorph microactuator to enable the creation of novel locomoting microrobots. 

The rest of the paper is organized as follows. \mbox{Section\,\ref{SEC02}} presents the design and fabrication of the introduced \mbox{$10$-mg} \mbox{SMA-based} bimorph microactuator, and the \mbox{$30$-mg} \mbox{FRISSHBot} swimmer developed to test and demonstrate actuation functionality and performance. \mbox{Section\,\ref{SEC03}} describes and discusses the characterization experiments performed to evaluate the actuator's output range and \mbox{frequency-bandwidth} of operation. \mbox{Section\,\ref{SEC04}} presents locomotion experiments performed using the \mbox{FRISSHBot} and analyses of the measured swimming performance in terms of forward speed. Last, \mbox{Section\,\ref{SEC05}} summarizes the research presented in the paper and discusses directions for future work.

\section{Design and Fabrication} 
\label{SEC02}
\vspace{-0.5ex}
The design and functionality of the proposed \mbox{SMA-based} bimorph actuator and \mbox{FRISSHBot} swimmer are graphically explained in~\mbox{Fig.\,\ref{FIG02}}. As seen in~\mbox{Fig.\,\ref{FIG02}(a)}, the swimmer is composed of four main elements: \mbox{(i)~an} \mbox{SMA-based} bimorph actuator that drives the propulsor; \mbox{(ii)~two} pads made of \mbox{\textit{carbon fiber}} (CF) that support the weight of the swimmer on water by leveraging surface tension; \mbox{(iii)~a} head that, interacting with water, generates high lateral drag; and, \mbox{(iv)~a} soft tail made of \mbox{$25$-{\textmu}m-thick} fluoropolymer film~\mbox{(AirTech~A4000R14417)}. Drawing from the designs in~\cite{SMALLBug_2020,SMARTI_2021,RoBeetle_2020,Waterstrider_2023,VLEIBot_2024}, the proposed bimorph actuator combines two SMA subsystems that can be excited independently to sequentially produce output displacements in two directions---e.g., up and down---during an actuation cycle. Consequently, as seen in~\mbox{Fig.\,\ref{FIG02}(b)}, the actuator can be operated according to two modes: \mbox{(i)~unimorph} actuation in one of the two possible directions of deformation, corresponding to periodic transitions between the relaxed state (\textit{detwinned martensite} phase) and sequentially one of the two fully deformed states (\textit{austenite} phase) as defined in~\mbox{Fig.\,\ref{FIG02}(b)}, and vice versa; and, \mbox{(ii)~bimorph} actuation, corresponding to direct periodic transitions between the two fully deformed states (austenite phase) as depicted in~\mbox{Fig.\,\ref{FIG02}(b)}. A key design element of the bimorph actuator that enables the functionality and high performance of the \mbox{FRISSHBot} is the installation of its driving SMA wires with a small angle $\alpha$ \mbox{($\sim\hspace{-0.3ex}3\,^{\circ}$)} and sufficient separation to avoid overlapping and collisions with the central beam of the device, as depicted in the inset of~\mbox{Fig\,\ref{FIG02}(a)}. 

In the course of the iterative design and development process of the actuator, experiments indicated that parallel placement of the SMA wires---using for example the methods described in~\mbox{\cite{SMALLBug_2020}~and~\cite{Waterstrider_2023}}---leads to mechanical failure of the SMA material due to excessive strain at one side of the actuator when the opposite side is contracted during an actuation cycle. As shown in the inset of~\mbox{Fig\,\ref{FIG02}(a)}, by avoiding contact between the SMA wires and central beam of the actuator, the wires deform linearly and no undesired forces are exerted on them during cyclic operation. Consequentially, the SMA material is not stressed and strained beyond its operational limits, and mechanical failure is prevented. The \mbox{FRISSHBot} locomotes forward by operating the driving actuator in the bimorph mode, as illustrated in~\mbox{Fig.\,\ref{FIG02}(b)}. Theoretically, this swimmer can perform turning maneuvers using the same techniques devised for the \mbox{VLEIBot} platform presented in~\cite{VLEIBot_2024}. Namely, by operating the actuator in either of the two possible directions of displacement according to the unimorph mode, right or left turns can be achieved. In contrast with the actuator that drives the \mbox{VLEIBot}, the proposed bimorph actuator can excite the \mbox{FRISSHBot} over a relatively wide frequency range (from $0$ to \mbox{$20$\,Hz}) without the need for an offline tuning procedure. 
\begin{figure}[t!]
\vspace{1.5ex}
\begin{center}
\includegraphics[width=0.48\textwidth]{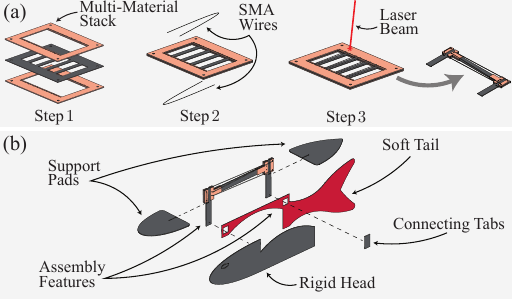}
\end{center}
\vspace{-2ex}
\caption{\textbf{Fabrication of the proposed \mbox{SMA-based} bimorph actuator and FRISSHBot.} \textbf{(a)}~The actuator is fabricated in three steps. In \mbox{Step\,1}, a \mbox{multi-material} stack composed of three premachined layers is aligned and fastened using four pins according to the technique described in~\cite{BeePlus_2019}, and then secured with CA glue; the top and bottom layers are made of \mbox{Cu-FR4} material, and the central layer is made of CF. In \mbox{Step\,2}, using the technique described in~\cite{SMALLBug_2020}, SMA \mbox{NiTi} wires are tied in tension, and secured using simple knots and drops of CA glue, to two opposite sides of the assembled stack over the CF structures that become the central \mbox{beam-springs} of the actuators after fabrication; the diameter and nominal transition temperature of the SMA wires are \mbox{$38.1$\,{\textmu}m} and \mbox{$90\,^{\circ}\ts{C}$}, respectively. In \mbox{Step\,3}, the actuators are released from the assembled \mbox{multi-material} stack using a \mbox{$3$-W~$355$-nm} DPSS laser~\mbox{(Photonics\,Industries\,DCH-$355$-$3$)}. \textbf{(b)}~The \mbox{FRISSHBot} is fabricated according to the following procedure. First, the soft tail is cut from fluoropolymer film~\mbox{(AirTech~A4000R14417)} and mechanically connected to the bimorph actuator using integrated assembly tabs. Then, the support pads, head, and connecting tabs are cut out of CF layers and assembled to create the swimmer. Last, we connect \mbox{$52$-AWG} tether wires to the SMA subsystems at each side of the bimorph actuator, using conductive silver epoxy \mbox{(MG\,Chemicals\,$8331$D)}. \label{FIG03}}
\vspace{-2ex}
\end{figure}

We empirically designed the head of the \mbox{FRISSHBot} using an iterative method with the objective of maximizing the generation of lateral drag during swimming; as seen in the accompanying supplementary movie, the resulting lateral drag produced by the swimmer's head is high relative to that produced by the soft tail of the propulsor. This feature was conceived to facilitate the \textit{anchoring} of the head, which is connected to one end of the actuator, in the surrounding fluid while the other end of the actuator \textit{flaps} the tail of the propulsor to generate thrust. The basic geometry of the propulsor's tail was loosely inspired by the planform of \textit{carangiform} swimmers and, then, heuristically improved through swimming experiments; additionally, we considered as a main design guideline the interchangeability of the tail and head of the prototype. Because of its main design characteristics, the \mbox{FRISSHBot} platform can be used to dynamically characterize the force production for locomotion of a wide gamut of different tail geometries and undulatory modes. The fabrication process of the proposed \mbox{$10$-mg} actuator is depicted in~\mbox{Fig.\,\ref{FIG03}(a)}. This process consists of three steps. In \mbox{Step\,1}, a \mbox{multi-material} stack composed of three premachined layers is aligned and fastened using four pins according to the technique described in~\cite{BeePlus_2019}, and then secured with \textit{cyanoacrylate}~(CA) glue; the top and bottom layers are made of \mbox{copper-clad} FR4~(\mbox{Cu-FR4}) material, and the central layer is made of CF. In \mbox{Step\,2}, using the technique described in~\cite{SMALLBug_2020}, SMA nitinol (\mbox{$56\,\%$\,Ni}, \mbox{$44\,\%$\,Ti}) wires are tied in tension, and secured using simple knots and drops of CA glue, to two opposite sides of the assembled stack over the CF structures that become the central \mbox{beam-springs} of the actuators after fabrication; the diameter and nominal transition temperature of the SMA wires are \mbox{$38.1$\,{\textmu}m} and \mbox{$90\,^{\circ}\ts{C}$}, respectively. In \mbox{Step\,3}, the actuators are released from the assembled \mbox{multi-material} stack using a \mbox{$3$-W~$355$-nm} \textit{\mbox{diode-pumped~solid-state}}~(DPSS) laser~\mbox{(Photonics\,Industries\,DCH-$355$-$3$)}. 

The assembly of a precursor stack can be intelligently devised to include geometrical features that facilitate the final integration of the actuator into the fabricated microrobot. In this case, we added CF posts that are used to attach the main components of the \mbox{FRISSHBot}. As seen in the inset of~\mbox{Fig.\,\ref{FIG02}(a)}, to ensure the electrical isolation of the two sides of the actuator's bimorph configuration, strips of the \mbox{Cu-FR4} layer are removed at each end of the device. Last, pairs of \mbox{$52$-AWG} tether power wires are connected to thin tiles of Cu at both sides of the actuator and secured with conductive silver epoxy~\mbox{(MG\,Chemicals\,$8331$D)}, employing the method described in~\cite{Waterstrider_2023}. The final assembly of a \mbox{FRISSHBot} prototype is depicted in~\mbox{Fig.\,\ref{FIG03}(b)}. As seen, the two support pads, head, and soft tail are connected to the actuator to physically realize the microrobot using the assembly features and connecting tabs indicated in the illustration. The use of tabs ensures a robust mechanical bonding between the CF pieces and fluoropolymer film. 
\begin{figure}[t!]
\vspace{1.5ex}    
\begin{center}
\includegraphics[width=0.48\textwidth]{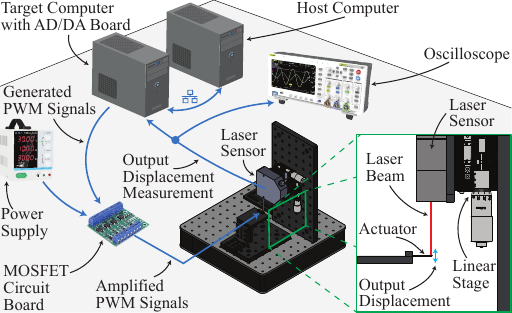}
\end{center}
\vspace{-2ex}
\caption{\textbf{Experimental setup used to characterize the proposed \mbox{$\bs{10}$-mg} \mbox{SMA-based} bimorph actuator for microrobotics.} To assess the functionality, performance, and operational range of the actuator, we employed a laser displacement sensor (Keyence \mbox{LK-G$32$}), a Mathworks Simulink \mbox{Real-Time} \mbox{host--target} system equipped with an \mbox{AD/DA} board (National Instruments \mbox{PCI-$6229$}), and a MOSFET circuit board (\mbox{four-channel}~\mbox{YYNMOS-$4$}). In this case, the \mbox{host--target} configuration generates the PWM signals that control two independent channels of the MOSFET circuit board used to excite the actuator. As seen, the power required by the MOSFET circuits to amplify the PWM signals is provided by an external \mbox{power-supply} unit. Essentially, the two PWM voltages outputted by the \mbox{AD/DA} board open and close the two channels of the MOSFET circuit board that function as switches for the power provided by the external \mbox{power-supply} unit that excite the SMA wires of the actuator. In this scheme, the SMA materials of the subsystems at both sides of the actuator are periodically Joule heated and allowed to cool down to induce transitions from the detwinned martensite phase to the austenite phase, and vice versa. As seen in the inset, during the performance of the characterization tests, one distal end of the bimorph actuator is held precisely aligned under the laser sensor, which outputs a voltage proportional to the actuator displacement. Also, we place the tested actuator at a distance from the sensor such that the zero output corresponds to the center of the measurement range. Last, the output displacement measurement is sampled and recorded by the \mbox{host--target} system using the \mbox{AD/DA} board of the target computer. All signals transmitted between the components of the experimental setup are generated and sampled at a frequency of \mbox{$2$\,kHz}. \label{FIG04}}
\vspace{-2ex}
\end{figure}
\begin{figure*} [t!]
\vspace{1.5ex}
\begin{center}    
\includegraphics[width=0.98\textwidth]{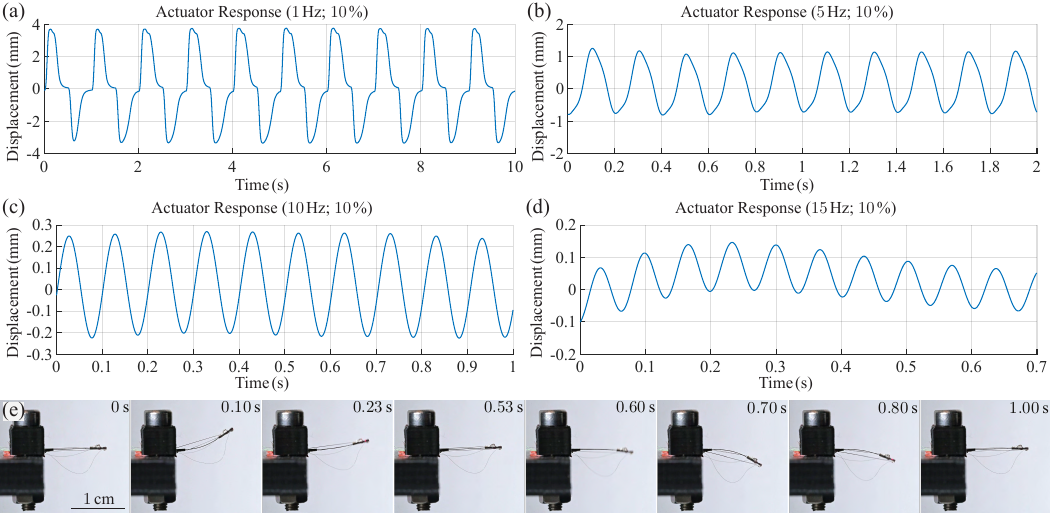}
\end{center}
\vspace{-2ex}
\caption{\textbf{Dynamic responses of the tested \mbox{SMA-based} bimorph actuator.} \textbf{(a)}~Ten cycles of the \mbox{steady-state} actuator response to a PWM excitation with frequency of \mbox{$1$\,Hz} and DC of $10\,\%$. In this test, we measured an AMADO value of \mbox{$7.08$\,mm}. \textbf{(b)}~Ten cycles of the \mbox{steady-state} actuator response to a PWM excitation with frequency of \mbox{$5$\,Hz} and DC of $10\,\%$. In this test, we measured an AMADO value of \mbox{$1.83$\,mm}. \textbf{(c)}~Ten cycles of the \mbox{steady-state} actuator response to a PWM excitation with frequency of \mbox{$10$\,Hz} and DC of $10\,\%$. In this test, we measured an AMADO value of \mbox{$0.56$\,mm}. \textbf{(d)}~Ten cycles of the \mbox{steady-state} actuator response to a PWM excitation with frequency of \mbox{$15$\,Hz} and DC of $10\,\%$. In this test, we measured an AMADO value of \mbox{$0.28$\,mm}. \textbf{(e)}~\mbox{One-second} photographic sequence of actuator response to a PWM excitation with frequency of \mbox{$1$\,Hz} and DC of \mbox{$10\,\%$}. Video footage of these experiments can be viewed in the accompanying supplementary movie. \label{FIG05}}
\vspace{-2ex}
\end{figure*}

\section{Characterization of the Bimorph Actuator}
\label{SEC03}
\vspace{-0.5ex}
\subsection{Experimental Setup} 
\label{SEC03A}
\vspace{-0.5ex}
To assess the functionality and performance of the proposed \mbox{SMA-based} bimorph actuator, we used the experimental setup depicted in~\mbox{Fig.\,\ref{FIG04}}. During tests, we generate, monitor, and record signals using a Mathworks Simulink \mbox{Real-Time} \mbox{host--target} system. The target computer is equipped with an \mbox{\textit{analog-digital/digital-analog}} \mbox{(AD/DA)} board (National~Instruments~\mbox{PCI-$6229$}), which generates the two PWM signals inputted to the MOSFET circuit board \mbox{(four-channel YYNMOS-$4$)} that provides the two currents necessary to Joule heat the two pieces of SMA material of the tested actuator. To consistently ensure a periodic sequential switching between the two directional states of the bimorph actuator (up and down), we generate two separate PWM signals with identical \mbox{\textit{duty cycle}}~(DC) and frequency, but with a phase shift of \mbox{$180^{\circ}$} between them, which ensures that both sides of the actuator are not excited simultaneously during operation.

Prior to performing characterization experiments, we mount the tested bimorph actuator on the stand shown in~\mbox{Fig.\,\ref{FIG04}}, where it is precisely aligned under a laser displacement sensor (\mbox{Keyence\,LK-G$32$}). The sensor measurement is read and recorded using the \mbox{host--target} system at a rate of \mbox{$2$\,kHz}. We use an oscilloscope to monitor the signals transmitted between the components of the setup and also precisely place the moving distal end of the tested actuator to obtain an instantaneous measurement in the range of the laser sensor. The voltages of both exciting PWM signals during the \textit{on} and \textit{off} sections of the PWM period are $15$ and \mbox{$0$\,V}, respectively. As usual, the ratio between the \textit{on} section and PWM period defines the PWM DC in per unit. An \mbox{\textit{on}-height} of \mbox{$15$\,V} ensures that the power circuits at both sides of the bimorph actuator are not excited with currents higher than \mbox{$250$\,mA} during the Joule heating of the SMA wires. Empirical evidence indicates that currents higher than \mbox{$250$\,mA} thermally damage the \mbox{$52$-AWG} tether wires that connect the actuator with the amplifying \mbox{MOSFET} circuit. The main experimental variable that we investigated in this research is the \mbox{MADO}---corresponding to the \mbox{peak-to-peak} displacement output in a single actuation period---in relation to its dependence on the frequency and DC of the exciting PWM signals. We tested the \mbox{frequency--DC}, $\left\{f,\ts{DC}\right\}$, pairs formed by combining the elements in the sets \mbox{$f \in \left\{ 1,5,10,15,20 \right\}$\,Hz} and \mbox{$\ts{DC} \in \left\{1\hspace{-0.3ex}:\hspace{-0.3ex}1\hspace{-0.3ex}:\hspace{-0.3ex}10\right\}$\,\%}. For each tested pair, we excited the bimorph actuator for \mbox{$30$\,s}. Due to the complex hysteretic dynamics of the \mbox{SMA-based} actuation mechanism, all tested actuators exhibited marked transient periods before reaching \mbox{steady-state} operation; for this reason, we computed the \textit{average} \mbox{MADO} (AMADO) from \mbox{$15$\,s} of \mbox{steady-state} data. After collecting the measured \mbox{output-displacement} data, we processed them using a digital \mbox{zero-phase} \textit{finite impulse response}~(FIR) \mbox{low-pass} filter with order $10^3$ and a cutoff frequency of \mbox{$100$\,Hz}, synthesized using the digital filter design Matlab tool.

\subsection{Results and Discussion} 
\label{SEC03B}
\vspace{-0.5ex}
\mbox{Steady-state} responses of the tested actuator to PWM excitations with a DC of \mbox{$10\,\%$} and frequencies of $1$, $5$, $10$, and \mbox{$15$\,Hz} are shown in~\mbox{Figs.\,\ref{FIG05}(a)--(d)}, respectively. Here, each plot shows $10$ actuation cycles of the corresponding response. At low frequencies (\mbox{$\sim\hspace{-0.3ex}1$\,Hz}), the shapes of the rising sides of the responses' cyclic oscillations resemble the form of a tangent function. In the transient responses excited by PWM signals with frequencies of $5$ and \mbox{$10$\,Hz} (not shown), it can be observed that the \mbox{one-cycle} MADO slowly increases over time until reaching steady state in about \mbox{$10$\,s}; also, as seen in \mbox{Figs.\,\ref{FIG05}(b)--(c)}, the shapes of the \mbox{steady-state} actuator responses become approximately sinusoidal. At frequencies of \mbox{$15$\,Hz} (see~\mbox{Fig.\,\ref{FIG05}(d)}) and higher, we continue to observe a sinusoidal pattern; however, we also observe a \mbox{low-frequency} component corresponding to an \mbox{output-displacement} bias that slowly and periodically fluctuates with time. One main characteristic and functional advantage of the bimorph actuator introduced here is that, for exciting frequencies lower than \mbox{$15$\,Hz}, the measured \mbox{steady-state} actuation biases are significantly smaller than those reported for \mbox{SMA-based} unimorph microactuators~\cite{SMALLBug_2020,Waterstrider_2023}. By decreasing the actuation bias, we effectively removed the need to account for it during the process of microrobotic design and fabrication, provided that the operation frequencies remain below \mbox{$15$\,Hz}.

\mbox{Fig.\,\ref{FIG05}(e)} shows a photographic sequence with the different stages of an actuation cycle in response to a \mbox{$1$-Hz} PWM excitation. As seen in the first frame (\mbox{$0$\,s}), the tested bimorph actuator starts at its relaxed state; next, as seen in the second frame (\mbox{$0.1$\,s}), the \textit{top} SMA wires are thermally excited---transitioning from martensite to austenite---and deform the actuator in the upward direction; next, as seen in the third and fourth frames (\mbox{$0.23$~and~$0.53$\,s}), the top SMA wires cool down---transitioning from austenite to martensite---and the actuator begins to return to its relaxed state; next, as seen in the fifth and sixth frames (\mbox{$0.6$~and~$0.7$\,s}), the \textit{bottom} SMA wires are thermally excited---transitioning from martensite to austenite---and deform the actuator in the downward direction; next, as seen in the seventh frame (\mbox{$0.8$\,s}), the bottom SMA wires cool down---transitioning from austenite to martensite---and the actuator begins to return to its relaxed state; last, as seen in the eighth frame (\mbox{$1.0$\,s}), the actuator returns to its relaxed state. Video footage of this and other experiments corresponding to PWM excitations with frequencies of $5$, $10$, $15$, and \mbox{$20$\,Hz} can be seen in the accompanying supplementary movie. The change in shape, as the exciting PWM frequency increases, of the instantaneous actuator displacement is explained by the intricate nonlinear dynamic relationships between the strain, stress, and temperature of the actuator's SMA wires during operation. In~\cite{JoeyMech2021}, we presented and discussed the hysteretic \mbox{strain--temperature} relationships of an SMA wire functioning under several different constant stress values. As discussed therein, the \mbox{strain--temperature} curves shift to the \mbox{upper-right} of the plots---i.e., toward higher temperature and higher strain---as the constant stress applied to the wire is increased. Thus, it follows that at higher stress values, the SMA wires transition from austenite to martensite, and vice versa, at higher temperatures. In the case of the proposed actuator, this phenomenon, in combination with adequate PWM excitation, can be leveraged to increase the speed and smoothness of operation.   

At low actuation frequencies (\mbox{$\sim\hspace{-0.3ex}1$\,Hz}), as in the case shown in~\mbox{Fig.\,\ref{FIG05}(a)}, the actuator returns from a deflected state to the relaxed state following an \mbox{exponential-resembling} trajectory before deflecting in the opposite direction, which indicates that the SMA material cools down significantly below its transition temperature during each actuation cycle. As seen in~\mbox{Figs.\,\ref{FIG05}(b)--(d)}, as the frequency of operation increases, the actuator's response appears to increasingly pass through the relaxed state more smoothly and directly. This phenomenon indicates that, during an actuation cycle, the top SMA wire does not cool down significantly below its transition temperature before the bottom SMA wire begins to contract; similarly, the bottom SMA wire does not cool down significantly below its transition temperature before the top SMA wire begins to contract. We deduce from these observations that, at higher frequencies of operation, the SMA wires during cooling experience an additional stress generated by the SMA wires at the opposite side of the actuator. This additional stress facilitates the transition of the SMA material from austenite to detwinned martensite---and elongation of the SMA wires---at higher temperatures. Therefore, given the bimorph configuration of the tested actuator, both SMA subsystems---top and bottom---function following relatively fast and smooth minor hysteretic loops, as defined in~\cite{RoBeetle_2020} and \cite{JoeyMech2021}. This phenomenon explains why the actuator's response becomes smoother and more sinusoidal over the range $5$ to \mbox{$10$\,Hz}, which indicates potentially excellent functionality in many microrobotic applications.
\begin{figure}[t!]
\vspace{1.5ex}
\begin{center}
\includegraphics[width=0.48\textwidth]{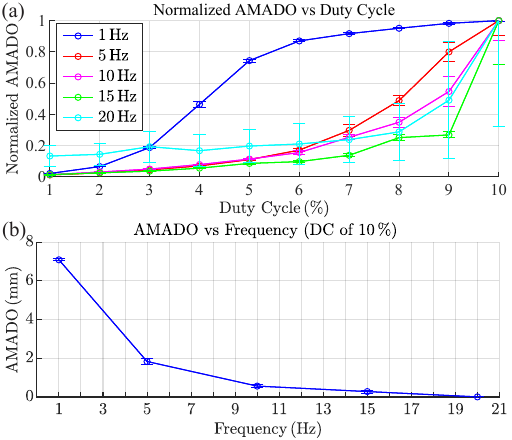}
\end{center}
\vspace{-2ex}
\caption{\textbf{Characterization of the tested \mbox{SMA-based} bimorph actuator.} \textbf{(a)}~Measured normalized AMADO values for exciting PWM voltages defined by the \mbox{frequency--DC}, $\left\{f,\ts{DC}\right\}$, pairs formed by combining the elements in the sets \mbox{$f\in\left\{1,5,10,15,20 \right\}$\,Hz} and \mbox{$\ts{DC}\in\left\{1\vspace{-0.3ex}:\vspace{-0.3ex}1\vspace{-0.3ex}:\vspace{-0.3ex}10 \right\}\,\%$}, computed from \mbox{$15$\,s} of \mbox{steady-state} experimental data. As seen, for all the tested PWM frequencies, the AMADO value steadily increases as the DC is increased. \textbf{(b)}~Measured AMADO values for PWM frequencies in the set \mbox{$f\in\left\{1,5,10,15,20 \right\}$\,Hz} and DC of \mbox{$10$\,\%}. As seen, the AMADO value exponentially decreases as the PWM frequency is increased. In this case, the measured AMADO values are $7.08$, $1.83$, $0.56$, $0.28$, and \mbox{$0.006$\,mm} for PWM frequencies of $1$, $5$, $10$, $15$, and \mbox{$20$\,Hz}, respectively. In both plots, the vertical bars indicate the calculated standard deviations. \label{FIG06}}
\vspace{-2ex}
\end{figure}

\mbox{Fig.\,\ref{FIG06}} presents experimental results on the characterization of the tested bimorph actuator in terms of the AMADO value. Each point in~\mbox{Fig.\,\ref{FIG06}(a)} shows the AMADO value corresponding to the excitation defined by the pair $\left\{ f,\ts{DC}\right\}$, with \mbox{$f \in \left\{1,5,10,15,20 \right\}$\,Hz} and \mbox{$\ts{DC} \in \left\{1\hspace{-0.3ex}:\hspace{-0.3ex}1\hspace{-0.3ex}:\hspace{-0.3ex}10\right\}$\,\%}, computed from \mbox{$15$\,s} of experimental data. Here, for each tested frequency, we normalized the AMADO values with respect to the maximum computed AMADO value among those corresponding to the ten tested DCs. The set of maximum AMADO values for the five tested frequencies is \mbox{$\ts{AV}_{\ts{max}} \in \left\{7.08,1.83,0.56,0.28,0.006\right\}$\,mm} (\mbox{see~Fig.\,\ref{FIG06}(b)}). In almost all the tested cases ($49$ out of $50$), the AMADO value markedly increases as the DC is increased up to \mbox{$10$\,\%}; empirical evidence indicates that, in dry air, PWM excitations with DCs greater than \mbox{$10$\,\%} might cause the tested actuator to get stuck in one actuation state and, therefore, were not included in the characterization tests discussed here. As clearly seen in~\mbox{Fig\,\ref{FIG06}(b)}, the maximum AMADO value decreases markedly as the actuation frequency increases. We hypothesize that it is possible to widen the actuation bandwidth by using \mbox{high-temperature} SMA materials and optimizing the geometry of the SMA wires to increase the associated cooling rates. In both plots of \mbox{Fig.\,\ref{FIG06}}, the vertical bars indicate the calculated standard deviations.

Overall, the presented \mbox{SMA-based} bimorph actuator---with a weight of \mbox{$10$\,mg}, volume of \mbox{$4.8\,\ts{mm}^3$}, and operational frequency of \mbox{$20$\,Hz}---is much lighter, smaller, and faster relative to its scale than any other \mbox{state-of-the-art} \mbox{SMA-based} bimorph actuator reported to date. For example, the actuators presented in~\cite{Geetha2022BimorphSwitch} and \cite{Chen2022BendingSMA} weigh more than \mbox{$300$\,mg}, have volumes greater than \mbox{$50\,\ts{mm}^3$}, and exhibit slow actuation responses (\mbox{$>1$\,s}); and, the actuators presented in~\cite{Kennedy2020BimorphSMA} exhibit excellent operational bandwidths (up to \mbox{$40$\,Hz}) and \mbox{low-frequency} output displacements (\mbox{$>25$\,mm}), but their volumes are on the order of \mbox{$48$~to~$96\,\ts{mm}^3$} and their average \mbox{power-consumption} estimates are on the order of \mbox{$0.25$~to~$0.75$\,W}, features that greatly limit their applicability to microrobotics. 

\begin{figure}
\vspace{1.5ex}    
\begin{center}    
\includegraphics[width=0.48\textwidth]{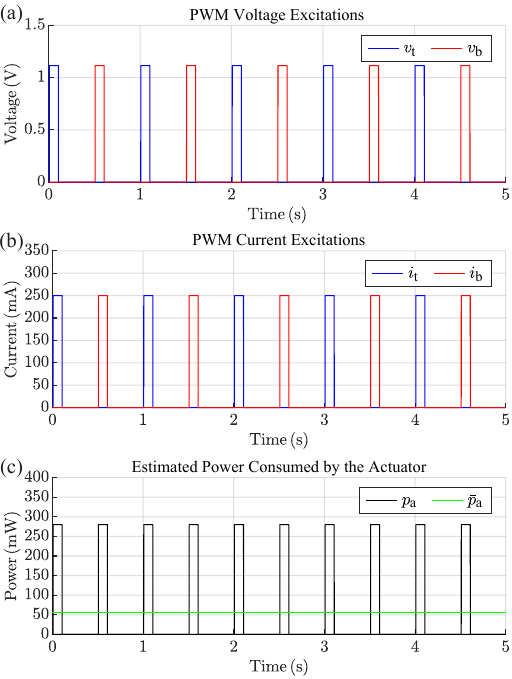}
\end{center}
\vspace{-2ex}
\caption{\textbf{Estimated power consumed by the tested actuator during operation at \mbox{$\bs{1}$\,Hz} and PWM DC of \mbox{$\bs{10}$\,\%}.} \textbf{(a)}~Theoretical PWM voltages, $v_{\ts{t}}$ and $v_{\ts{b}}$, that excite the top and bottom SMA subsystems of the bimorph actuator, respectively. As seen, there exists a \mbox{$180\,^{\circ}$} phase shift between these two PWM voltages and their \mbox{\textit{on}-height} is \mbox{$1.11$\,V}. \textbf{(b)}~Theoretical PWM currents, $i_{\ts{t}}$ and $i_{\ts{b}}$, which according to Ohm's law result from independently Joule heating the top and bottom SMA subsystems of the bimorph actuator with the voltages $v_{\ts{t}}$ and $v_{\ts{b}}$ \mbox{in~(a)}. We calculated the equivalent resistance of each SMA subsystem (top and bottom), \mbox{$r_{\ts{t}} = r_{\ts{b}} = r_{\ts{a}}=4.45\,{\Omega}$}, using the data provided by the manufacturer (Dynalloy) regarding the properties of the SMA material. We chose the excitations $v_{\ts{t}}$ and $v_{\ts{b}}$ such that the maximum value of both $i_{\ts{t}}$ and $i_{\ts{b}}$ is \mbox{$250$\,mA}, which was found experimentally to be the maximum current that the \mbox{$52$-AWG} tether wires used in the characterization experiments can withstand. \textbf{(c)}~Total theoretical instantaneous and average power consumption of
the tested \mbox{SMA-based} bimorph actuator during operation, $p_{\ts{a}}$ and $\bar{p}_{\ts{a}}$, corresponding to the PWM excitations in (a). As seen, the maximum total instantaneous power consumed by the tested actuator is \mbox{$280$\,mW}, which occurs when either side of the device is active. Consistently, the estimated average consumed power during operation is \mbox{$56$\,mW}. In this estimation, we did not consider the power loss as heat through the tether wires because they are not required for delivering onboard battery power to autonomous microrobots. \label{FIG07}}
\vspace{-2ex}
\end{figure}

A relevant dynamic characteristic of any robotic device is power consumption; in the case of \mbox{mm-scale} actuators, it fundamentally determines the amount of energy a microrobot must carry onboard to achieve autonomy for sufficiently long periods of time to complete tasks in unstructured environments. Here, we assess the power requirements of the proposed bimorph actuator using \mbox{first-principles} analyses informed by experimental data. In the estimation, we do not include the energy dissipated by the \mbox{$52$-AWG} tether wires used in the characterization experiments because they are not required when this type of actuator is employed to drive autonomous microrobotic platforms. Nominally, each SMA subsystem at each side of the actuator is excited with a current of \mbox{$250$\,mA} when the applied PWM voltage is in the \textit{on} state. We calculated the theoretical resistances of the SMA subsystems using the geometry of the actuator and property data of the SMA material provided by the manufacturer (Dynalloy). In this case, the estimated value for the total resistance of each SMA \mbox{sub-circuit} is \mbox{$r_{\ts{a}}=4.45\,\Omega$}. Thus, the instantaneous power consumed by the actuator in steady state, $p_{\ts{a}}$, can simply be estimated as
\begin{align}
p_{\ts{a}} = v_{\ts{t}} i_{\ts{t}} + v_{\ts{b}} i_{\ts{b}} =
i_{\ts{t}}^2r_{\ts{t}} + i_{\ts{b}}^2r_{\ts{b}} =
\left(i_{\ts{t}}^2 + i_{\ts{b}}^2\right) r_{\ts{a}},
\label{EQN01}
\end{align}
where $v_{\ts{t}}$, $i_{\ts{t}}$, and $r_{\ts{t}}$ are the voltage, current, and resistance corresponding to the top SMA subsystem; and, $v_{\ts{b}}$, $i_{\ts{b}}$, and $r_{\ts{b}}$ are the voltage, current, and resistance corresponding to the bottom SMA subsystem. By design, we know that \mbox{$r_{\ts{t}} = r_{\ts{b}} = r_{\ts{a}}$}. \mbox{Fig.\,\ref{FIG07}(a)} shows the theoretical $v_{\ts{t}}$ (blue) and $v_{\ts{b}}$ (red) used in the estimation of power consumption. These voltages are PWM signals with a frequency of \mbox{$1$\,Hz}, DC of \mbox{$10$\,\%}, and \mbox{\textit{on}-height} of \mbox{$1.1$\,V}; the corresponding $i_{\ts{t}}$ (blue) and $i_{\ts{b}}$ (red) are shown in~\mbox{Fig.\,\ref{FIG07}(b)}; and, the computed $p_{\ts{a}}$ and average consumed power, $\bar{p}_{\ts{a}}$, are shown in~\mbox{Fig.\,\ref{FIG07}(c)}. As seen, the estimated peak and average of power consumption are $280$ and \mbox{$56$\,mW}, respectively. These simple calculations indicate the suitability of the proposed \mbox{$10$-mg} bimorph actuator for autonomous microrobotic applications as it is physically feasible to obtain the estimated required power and voltage by employing \mbox{state-of-the-art} commercial \mbox{lithium-ion} batteries and simple power electronics.

\section{The FRISSHBot and Locomotion Experiments} 
\label{SEC04}
\vspace{-0.5ex}
To test and demonstrate the functionality and performance of the presented \mbox{$10$-mg} \mbox{SMA-based} bimorph actuator for microrobotic applications, we developed the \mbox{$30$-mg} \mbox{FRISSHBot} swimmer, already described in Section\,\ref{SEC02} and shown in~\mbox{Fig.\,\ref{FIG01}}. The \mbox{FRISSHBot} is propelled by a \mbox{forked-caudal-shaped} soft tail and a rigid CF head; we selected the final shapes of both the tail and head through an iterative heuristic process, according to which we performed locomotion experiments and considered forward speed as the figure of merit for empirical optimization. Experimental results indicate that to maximize both the forward force and speed generated by the FRISSHBot, the rigid head must generate a lateral drag significantly higher than that produced by the undulating soft tail. This phenomenon suggests that hydrodynamic interactions between the flows excited by the head and tail cause the forward thrust that produces locomotion. Currently, we cannot provide a thorough mathematical description of these interactions; however, we speculate that a high lateral drag inhibits large rotations of the swimmer's head, thus facilitating large oscillations of the tail relative to the swimmer's body and surrounding fluid. Also, we hypothesize that the rigid edge of the head produces \mbox{leading-edge} vortices responsible for additional forward thrust.
\begin{figure*}[t!]
\vspace{1.5ex}    
\begin{center}
\includegraphics[width=0.98\textwidth]{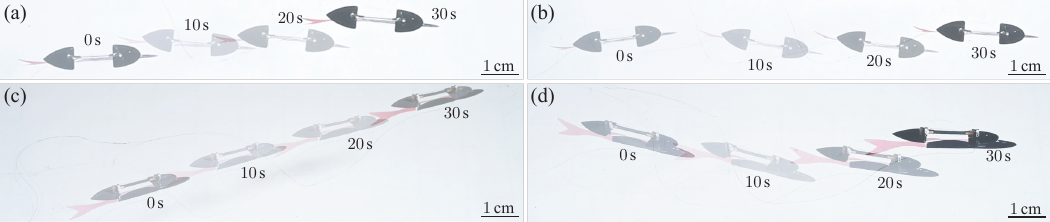}
\end{center}
\vspace{-2ex}
\caption{\textbf{Swimming experiments of the FRISSHBot.} \textbf{(a)}~Photographic composite of \mbox{top-view} frames---taken at \mbox{$10$-s} intervals---showing the FRISSHBot swimming excited by PWM voltages with a frequency of \mbox{$3$\,Hz} and DC of \mbox{$12$\,\%}. In this case, we measured a forward speed of about \mbox{$2.39\,\ts{mm}\cdot\ts{s}^{-1}$}~(\mbox{$0.07\,\ts{Bl}\cdot\ts{s}^{-1}$}). \textbf{(b)}~Photographic composite of \mbox{top-view} frames---taken at \mbox{$10$-s} intervals---showing the FRISSHBot swimming excited by PWM voltages with a frequency of \mbox{$4$\,Hz} and DC of \mbox{$12$\,\%}. In this case, we measured a forward speed of about \mbox{$3.06\,\ts{mm}\cdot\ts{s}^{-1}$}~(\mbox{$0.09\,\ts{Bl}\cdot\ts{s}^{-1}$}). \textbf{(c)}~Photographic composite of \mbox{side-view} frames---taken at \mbox{$10$-s} intervals---showing the FRISSHBot swimming excited by PWM voltages with a frequency of \mbox{$3$\,Hz} and DC of \mbox{$12$\,\%}. \textbf{(d)}~Photographic composite of \mbox{side-view} frames---taken at \mbox{$10$-s} intervals---showing the FRISSHBot swimming excited by PWM voltages with a frequency of \mbox{$4$\,Hz} and DC of \mbox{$12$\,\%}. Video footage of these swimming experiments can be viewed in the accompanying supplementary movie. \label{FIG08}}
\vspace{-2ex}
\end{figure*}

During the performance of the locomotion experiments discussed here, we excited the \mbox{FRISSHBot} prototype's bimorph actuator with \mbox{open-loop} PWM voltages generated using the same \mbox{real-time} \mbox{host--target} system described in~\mbox{Section\,\ref{SEC03A}} and depicted in~\mbox{Fig.\,\ref{FIG04}}. We selected the DCs and \mbox{\textit{on}-heights} of the exciting PWM voltages using the experimental results presented in~\mbox{Section\,\ref{SEC03B}}; specifically, \mbox{$12$\,\%} and \mbox{$15$\,V}, respectively. 
In this case, we increased the PWM DC value to compensate for the higher \mbox{heat-transfer} rates experienced by the actuator's SMA wires near the surface of water. 
Through the iterative \mbox{design-and-optimization} process described above, we discovered that the \mbox{FRISSHBot} achieves its best locomotion performance in terms of forward speed at excitation frequencies on the range from~$3$~to~\mbox{$4$\,Hz}. \mbox{Fig.\,\ref{FIG08}} shows photographic composites, made from frames taken at intervals of \mbox{$10$\,s}, of \mbox{$30$-s} swimming experiments when the FRISSHBot prototype is excited with these two frequencies. The top sequences in~\mbox{Figs.\,\ref{FIG08}(a)--(b)} correspond to top views and the bottom sequences in~\mbox{Figs.\,\ref{FIG08}(c)--(d)} correspond to side views. In the \mbox{$3$-Hz} case, we measured a forward speed of about \mbox{$2.39\,\ts{mm}\cdot\ts{s}^{-1}$~($0.07\,\ts{Bl}\cdot\ts{s}^{-1}$)}, which corresponds to a Reynolds number on the order of $80$. In the \mbox{$4$-Hz} case, we measured a forward speed of about \mbox{$3.06\,\ts{mm}\cdot\ts{s}^{-1}$~($0.09\,\ts{Bl}\cdot\ts{s}^{-1}$)}, which corresponds to a Reynolds number on the order of $100$. Video footage of these swimming experiments can be seen in the accompanying supplementary movie. While the swimming performance of the \mbox{FRISSHBot} can still be improved, the experimental results presented here clearly demonstrate that new types of microrobots---aquatic and terrestrial---can be developed using the proposed \mbox{$10$-mg} \mbox{SMA-based} bimorph actuator. To our best knowledge, the \mbox{FRISSHBot}---with a length of \mbox{$34$\,mm}, volume of \mbox{$24\,\ts{mm}^3$}, and mass of \mbox{$30$\,mg}---is the smallest and lightest microswimmer with onboard actuation developed to date. 

\section{Conclusion} 
\label{SEC05}
\vspace{-0.5ex}
We presented a new fully functional \mbox{$10$-mg} fast \mbox{SMA-based} bimorph actuator for microrobotics. This is the lightest and smallest microactuator of this type reported to date. This device can operate at frequencies of up to \mbox{$20$\,Hz} and produce \mbox{low-frequency} (\mbox{$\sim\hspace{-0.3ex}1$\,Hz}) output displacements on the order of $3.5$ and \mbox{$7.0$\,mm} in the unimorph and bimorph modes, respectively. To assess the operational range, functionality, and performance of the actuator, we performed a series of characterization experiments to determine the relationship between the DC and frequency of the exciting PWM voltage, and the measured AMADO value. Also, we estimated the actuator's power consumption during operation, and determined the suitability and feasibility of the proposed approach for autonomous microrobotic applications. To further test and demonstrate the functionality and suitability of the proposed bimorph actuator for microrobotics, we developed the \mbox{FRISSHBot}, a \mbox{$30$\,mg} microswimmer composed of a rigid head and a soft tail loosely inspired by the morphology of carangiform swimmers. 

In the \mbox{FRISSHBot's} design, the new \mbox{$10$-mg} bimorph actuator not only drives the swimmer but also serves as its main structure. After an \mbox{iterative--heuristic} optimization process, the resulting \mbox{FRISSHBot} prototype was capable of swimming at forward speeds as high as \mbox{$3.06\,\ts{mm}\cdot\ts{s}^{-1}$}~(\mbox{$0.09\,\ts{Bl}\cdot\ts{s}^{-1}$}). We hypothesize that hydrodynamic interactions between the flows generated by the head and tail of the \mbox{FRISSHBot} produce the thrust required for forward propulsion. This issue is a matter of current and further research in our laboratory. To our best knowledge, the \mbox{FRISSHBot} is the smallest and lightest fully functional microswimmer developed to date. While the locomotion performance achieved by the \mbox{FRISSHBot} can still be improved, the results presented here provide compelling evidence of the applicability of the presented approach to 
 \mbox{SMA-based} bimorph microactuation.

\bibliographystyle{IEEEtran}
\bibliography{references}

\begin{thebibliography}{10}
\providecommand{\url}[1]{#1}
\csname url@samestyle\endcsname
\providecommand{\newblock}{\relax}
\providecommand{\bibinfo}[2]{#2}
\providecommand{\BIBentrySTDinterwordspacing}{\spaceskip=0pt\relax}
\providecommand{\BIBentryALTinterwordstretchfactor}{4}
\providecommand{\BIBentryALTinterwordspacing}{\spaceskip=\fontdimen2\font plus
\BIBentryALTinterwordstretchfactor\fontdimen3\font minus \fontdimen4\font\relax}
\providecommand{\BIBforeignlanguage}[2]{{%
\expandafter\ifx\csname l@#1\endcsname\relax
\typeout{** WARNING: IEEEtran.bst: No hyphenation pattern has been}%
\typeout{** loaded for the language `#1'. Using the pattern for}%
\typeout{** the default language instead.}%
\else
\language=\csname l@#1\endcsname
\fi
#2}}
\providecommand{\BIBdecl}{\relax}
\BIBdecl

\bibitem{song2007PZTWaterStrider}
Y.~S. Song and M.~Sitti, ``{Surface-Tension-Driven Biologically Inspired Water Strider Robots: Theory and Experiments},'' \emph{IEEE Trans. Robot.}, vol.~23, no.~3, pp. 578--589, Jun. 2007.

\bibitem{lee2011design}
D.~Lee, S.~Kim, Y.-L. Park, and R.~J. Wood, ``{Design of \mbox{Centimeter-Scale} Inchworm Robots With Bidirectional Claws},'' in \emph{Proc. IEEE Int. Conf. Robot. Automat. (ICRA)}, Shanghai, China, May 2011, pp. 3197--3204.

\bibitem{goldberg2018power}
B.~Goldberg, R.~Zufferey, N.~Doshi, E.~F. Helbling, G.~Whittredge, M.~Kovac, and R.~J. Wood, ``{Power and Control Autonomy for High-Speed Locomotion With an Insect-Scale Legged Robot},'' \emph{IEEE Robot. Automat. Lett.}, vol.~3, no.~2, pp. 987--993, Apr. 2018.

\bibitem{wu2019insect}
Y.~Wu, J.~K. Yim, J.~Liang, Z.~Shao, M.~Qi, J.~Zhong, Z.~Luo, X.~Yan, M.~Zhang, X.~Wang, R.~S. Fearing, R.~J. Full, and L.~Lin, ``{\mbox{Insect-scale} fast moving and ultrarobust soft robot},'' \emph{Sci. Robot.}, vol.~4, no.~32, Jul. 2019, {Art.} no. eaax1594.

\bibitem{gravish2020stcrawler}
W.~Zhou and N.~Gravish, ``{Soft Microrobotic Transmissions Enable Rapid Ground-Based Locomotion},'' in \emph{Proc. IEEE/RSJ Int. Conf. Intell. Robots Syst. (IROS)}, Las Vegas, NV, USA, Oct. 2020, pp. 7874--7880.

\bibitem{BeePlus_2019}
X.~Yang, Y.~Chen, L.~Chang, A.~A. Calder\'on, and \mbox{N. O.} \mbox{P\'erez-Arancibia}, ``{Bee\textsuperscript{+}: A 95-mg Four-Winged Insect-Scale Flying Robot Driven by Twinned Unimorph Actuators},'' \emph{IEEE Robot. Automat. Lett.}, vol.~4, no.~4, pp. 4270--4277, Oct. 2019.

\bibitem{BeePlusPlus_2023}
R.~M. Bena, X.~Yang, A.~A. Calder\'on, and N.~O. \mbox{P\'erez-Arancibia}, ``{High-Performance Six-DOF Flight Control of the Bee\textsuperscript{++}: An Inclined-Stroke-Plane Approach},'' \emph{IEEE Trans. Robot.}, vol.~39, no.~2, pp. 1668--1684, Apr. 2023.

\bibitem{Li2023PZT}
K.~Li, X.~Zhou, Y.~Liu, J.~Sun, X.~Tian, H.~Zheng, L.~Zhang, J.~Deng, J.~Liu, W.~Chen, and J.~Zhao, ``{A 5 cm-Scale Piezoelectric Jetting Agile Underwater Robot},'' \emph{Adv. Intell. Syst.}, vol.~5, no.~4, Apr. 2023, {Art.} no 2200262.

\bibitem{contreras2017first}
D.~S. Contreras, D.~S. Drew, and K.~S.~J. Pister, ``{First Steps of a Millimeter-Scale Walking Silicon Robot},'' in \emph{Proc. 19th Int. Conf. Solid-State Sens. Actuators Microsyst. (TRANSDUCERS)}, Kaohsiung, Taiwan, Jun. 2017, pp. 910--913.

\bibitem{lu2018bioinspired}
H.~Lu, M.~Zhang, Y.~Yang, Q.~Huang, T.~Fukuda, Z.~Wang, and Y.~Shen, ``{A bioinspired multilegged soft millirobot that functions in both dry and wet conditions},'' \emph{Nat. Commun.}, vol.~9, Sep. 2018, {Art.} no. 3944.

\bibitem{hu2018small}
W.~Hu, G.~Z. Lum, M.~Mastrangeli, and M.~Sitti, ``{Small-scale soft-bodied robot with multimodal locomotion},'' \emph{Nature}, vol. 554, pp. 81--85, Feb. 2018.

\bibitem{pierre20183d}
R.~{St. Pierre}, W.~Gosrich, and S.~Bergbreiter, ``{A 3D-Printed \mbox{1\,mg} Legged Microrobot Running at 15 Body Lengths per Second},'' in \emph{Proc. Solid-State Sens. Actuators Microsyst. Workshop}, Hilton Head Island, SC, USA, Jun. 2018, pp. 59--62.

\bibitem{Duduta2017DEA}
M.~Duduta, D.~R. Clarke, and R.~J. Wood, ``{A High Speed Soft Robot Based On Dielectric Elastomer Actuators},'' in \emph{Proc. IEEE Int. Conf. Robot. Autom. (ICRA)}, Singapore, May 2017, pp. 4346--4351.

\bibitem{ren2022DEAFlyer}
Z.~Ren, S.~Kim, X.~Ji, W.~Zhu, F.~Niroui, J.~Kong, and Y.~Chen, ``{A High-Lift Micro-Aerial-Robot Powered by Low-Voltage and Long-Endurance Dielectric Elastomer Actuators},'' \emph{Adv. Mat.}, vol.~34, no.~7, Feb. 2022, {Art.} no. 2106757.

\bibitem{Berlinger2018DEAFish}
F.~Berlinger, M.~Duduta, H.~Gloria, D.~Clarke, R.~Nagpal, and R.~Wood, ``{A Modular Dielectric Elastomer Actuator to Drive Miniature Autonomous Underwater Vehicles},'' in \emph{Proc. IEEE Int. Conf. Robot. Autom. (ICRA)}, Brisbane, Australia, May 2018, pp. 3429--3435.

\bibitem{Ji2019DEAnsect}
X.~Ji, X.~Liu, V.~Cacucciolo, M.~Imboden, Y.~Civet, A.~E. Haitami, S.~Cantin, Y.~Perriard, and H.~Shea, ``{An autonomous untethered fast soft robotic insect driven by low-voltage dielectric elastomer actuators},'' \emph{Sci. Robot.}, vol.~4, no.~37, Dec. 2019, {Art.} no. eaaz6451.

\bibitem{MilliMobile2023}
K.~Johnson, Z.~Englehardt, V.~Arroyos, D.~Yin, S.~Patel, and V.~Iyer, ``{MilliMobile: An Autonomous Battery-free Wireless Microrobot},'' in \emph{Proc. 29th Annu. Int. Conf. Mob. Comput. Netw. (MobiCom)}, Madrid, Spain, Oct. 2023, pp. 1360--1375.

\bibitem{hutama2021UltrasonicMotor}
R.~Y. Hutama, M.~M. Khalil, and T.~Mashimo, ``{A Millimeter-Scale Rolling Microrobot Driven by a Micro-Geared Ultrasonic Motor},'' \emph{IEEE Robot. Automat. Lett.}, vol.~6, no.~4, pp. 8158--8164, Oct. 2021.

\bibitem{Truong2019JumpingRobot}
N.~T. Truong, H.~V. Phan, and H.~C. Park, ``{Design and demonstration of a bio-inspired flapping-wing-assisted jumping robot},'' \emph{Bioinspir. Biomim.}, vol.~14, no.~3, May 2019, {Art.} no. 036010.

\bibitem{Chen2020BioinspiredFlapper}
S.~Chen, L.~Wang, S.~Guo, C.~Zhao, and M.~Tong, ``{A Bio-Inspired Flapping Wing Robot of Variant Frequency Driven by Ultrasonic Motor},'' \emph{Appl. Sci.}, vol.~10, no.~1, Jan. 2020, {Art.} no. 412.

\bibitem{SMALLBug_2020}
X.-T. Nguyen, A.~A. Calder\'on, A.~Rigo, J.~Z. Ge, and N.~O. \mbox{P\'erez-Arancibia}, ``{SMALLBug: A 30-mg Crawling Robot Driven by a High-Frequency Flexible SMA Microactuator},'' \emph{IEEE Robot. Automat. Lett.}, vol.~5, no.~4, pp. 6796--6803, Oct. 2020.

\bibitem{SMARTI_2021}
R.~M. Bena, X.-T. Nguyen, A.~A. Calder\'on, and N.~O. \mbox{P\'erez-Arancibia}, ``{SMARTI: A 60-mg Steerable Robot Driven by High-Frequency Shape-Memory Alloy Actuation},'' \emph{IEEE Robot. Automat. Lett.}, vol.~6, no.~4, pp. 8173--8180, Oct. 2021.

\bibitem{RoBeetle_2020}
X.~Yang, L.~Chang, and N.~O. \mbox{P\'erez-Arancibia}, ``{An 88-milligram insect-scale autonomous crawling robot driven by a catalytic artificial muscle},'' \emph{Sci. Robot.}, vol.~5, no.~45, Aug. 2020, {Art.} no. eaba0015.

\bibitem{Waterstrider_2023}
C.~K. Trygstad, X.-T. Nguyen, and N.~O. \mbox{P\'erez-Arancibia}, ``{A New \mbox{1-mg} Fast Unimorph SMA-Based Actuator for Microrobotics},'' in \emph{Proc. IEEE/RSJ Int. Conf. on Intel. Robot. and Syst. (IROS)}, Detroit, MI, USA, Oct. 2023, pp. 2693--2700.

\bibitem{VLEIBot_2024}
E.~K. Blankenship, C.~K. Trygstad, F.~M. F.~R. Gon\c{c}alves, and N.~O. \mbox{P\'erez-Arancibia}, ``{VLEIBot: A New 45-mg Swimming Microrobot Driven by a Bioinspired Anguilliform Propulsor},'' in \emph{Proc. IEEE Int. Conf. Robot. Autom. (ICRA)}, Yokohama, Japan, May 2024, pp. 6014--6021.

\bibitem{Knick2019MEMSBimorph}
C.~R. Knick, D.~J. Sharar, A.~A. Wilson, G.~L. Smith, C.~J. Morris, and H.~A. Bruck, ``{High frequency, low power, electrically actuated shape memory alloy MEMS bimorph thermal actuators},'' \emph{J. Micromech. Microeng.}, vol.~29, no.~7, Jul. 2019, {Art.} no. 075005.

\bibitem{Sun2020Liftoff}
H.~Sun, J.~Luo, Z.~Ren, M.~Lu, D.~Nykypanchuk, S.~Mangla, and Y.~Shi, ``{Shape Memory Alloy Bimorph Microactuators by Lift-Off Process},'' \emph{J. Micro Nano-Manuf.}, vol.~8, no.~3, Sep. 2020, {Art.} no. 031003.

\bibitem{Seigner2021OrigamiBimorph}
L.~Seigner, G.~K. Tshikwand, F.~Wendler, and M.~Kohl, ``{Bi-Directional Origami-Inspired SMA Folding Microactuator},'' \emph{Actuators}, vol.~10, no.~8, Aug. 2021, {Art.} no. 181.

\bibitem{Geetha2022BimorphSwitch}
M.~Geetha, K.~Dhanalakshmi, and V.~Vetriselvi, ``{A shape memory alloy bimorph-actuated switch for antenna reconfiguration},'' \emph{J. Mater. Sci: Mater. Electron.}, vol.~33, pp. 4426--4437, Jan. 2022.

\bibitem{Chen2022BendingSMA}
J.~Chen, Q.~Ding, Y.~Kim, and S.~S. Cheng, ``{Design, modeling and evaluation of a millimeter-scale SMA bending actuator with variable length},'' \emph{J. Intell. Mater. Syst. Struct.}, vol.~33, no.~7, pp. 942--957, Apr. 2022.

\bibitem{Kennedy2020BimorphSMA}
S.~Kennedy, M.~Price, M.~Zabala, and E.~Perkins, ``{Vibratory Response Characteristics of High-Frequency Shape Memory Alloy Actuators},'' \emph{J. Vib. Acoust.}, vol. 142, no.~1, Feb. 2020, {Art.} no. 011004.

\bibitem{Cho2008SMARibbon}
K.-J. Cho, E.~Hawkes, C.~Quinn, and R.~J. Wood, ``{Design, fabrication and analysis of a body-caudal fin propulsion system for a microrobotic fish},'' in \emph{Proc. IEEE Int. Conf. Robot. Autom. (ICRA)}, Pasadena, CA, USA, May 2008, pp. 706--711.

\bibitem{Rossi2011BendingSMA}
C.~Rossi, J.~Colorado, W.~Coral, and A.~Barrientos, ``{Bending continuous structures with SMAs: a novel robotic fish design},'' \emph{Bioinsp. Biomim.}, vol.~6, no.~4, Dec. 2011, {Art.} no 045005.

\bibitem{JoeyMech2021}
J.~Z. Ge, L.~Chang, and N.~O. \mbox{P\'erez-Arancibia}, ``{\mbox{Preisach-model-based} position control of a shape-memory alloy linear actuator in the presence of time-varying stress},'' \emph{Mechatronics}, vol.~73, Feb. 2021, {Art.} no. 102452.

\end{thebibliography}
\end{document}